\documentclass[10pt,twocolumn,letterpaper]{article}

\usepackage{cvpr}              

\usepackage{graphicx}
\usepackage{amsmath}
\usepackage{amssymb}
\usepackage{booktabs}
\usepackage{multirow}
\usepackage{algorithm}
\usepackage{algorithmic}

\usepackage[pagebackref,breaklinks,colorlinks]{hyperref}

\usepackage[capitalize]{cleveref}
\crefname{section}{Sec.}{Secs.}
\Crefname{section}{Section}{Sections}
\Crefname{table}{Table}{Tables}
\crefname{table}{Tab.}{Tabs.}

\begin{document}
	
	\title{ An Effective Crop-Paste Pipeline for Few-shot Object Detection}

\author{Shaobo Lin, Kun Wang, Xingyu Zeng, Rui Zhao\\
	Sensetime Research\\
	{\tt\small $\{$linshaobo,wangkun,zengxingyu,zhaorui$\}$@sensetime.com}
}
	


\maketitle

\begin{abstract}
	Few-shot object detection (FSOD) aims to expand an object detector for novel categories given only a few instances for training. However, detecting novel categories with only a few samples usually leads to the problem of misclassification. In FSOD, we notice the false positive (FP) of novel categories is prominent, in which the base categories are often recognized as novel ones. To address this issue, a novel data augmentation pipeline that Crops the Novel instances and Pastes them on the selected Base images, called CNPB, is proposed. There are two key questions to be answered: (1) How to select useful base images? and (2) How to combine novel and base data? 
	We design a multi-step selection strategy to find useful base data. Specifically, we first discover the base images which contain the FP of novel categories and select a certain amount of samples from them for the 
	base and novel categories balance. Then the bad cases, such as the base images that have unlabeled ground truth or easily confused base instances, are removed by using CLIP. Finally, the same category strategy is adopted, in which a novel instance with category n is pasted on the base image with the FP of n.  
	During combination, a novel instance is cropped and randomly down-sized, and thus pasted at the assigned optimal location from the randomly generated candidates in a selected base image.  
	Our method is simple yet effective and can be easy to plug into existing FSOD methods, demonstrating significant potential for use. Extensive experiments on 
	PASCAL VOC and MS COCO validate the effectiveness of our method.  
\end{abstract}

\section{Introduction}
\label{sec:intro}

In recent years, we have witnessed the great progress of object detection~\cite{ren2016faster,redmon2017yolo9000,lin2017feature,carion2020end,dai2021dynamic}. However, the impressive performance of these models relies on a large amount of annotated data. The detectors cannot generalize well to novel categories, especially when the annotated data are scarce.  In contrast, humans can learn to recognize or detect a novel object with only a few labeled examples. Few-shot object detection (FSOD), which simulates this way, has attracted increasing attention. In FSOD, an object detector that is trained using base categories with sufﬁcient data (base images) can learn to detect novel categories using only a few annotated data (novel images).


\begin{figure}[t]
	\vskip -0.1in
	\begin{center}
		\centerline{\includegraphics[width=\columnwidth]{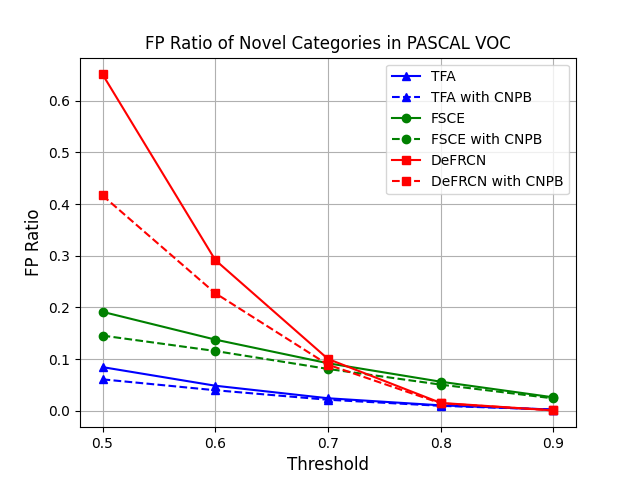}}
		\caption{The FP ratio of novel categories in the base dataset of PASCAL VOC split 1. We test TFA, FSCE and DeFRCN, which are three representative FSOD methods. The reported FP ratio is the average of multiple few-shot settings, including 1-shot, 3-shot, 5-shot and 10-shot. The threshold is used to count FP images whose confidence score on a novel category is higher than the threshold. Using our CNPB, the FP ratio of all methods decreases up to 20$\%$.}
		\label{fig0}
	\end{center}
	\vskip -0.3in
\end{figure}

The accuracy of a FSOD model is decided by the true positive (TP) and false positive (FP) of novel categories, since the novel categories are the main concern. 
The main problem of FSOD is misclassification~\cite{sun2021fsce}, in which novel categories are often recognized as base categories (the FP of base categories). However, the FP of novel categories is previously ignored, which is more important in FSOD. The FP of novel categories means the base categories and background regions that are recognized as the novel categories. 

We analyze the FP ratio of novel categories by using three representative FSOD methods, TFA~\cite{wang2020frustratingly}, FSCE~\cite{sun2021fsce}, and DeFRCN~\cite{qiao2021defrcn}, as shown in Figure~\ref{fig0}. We test the trained few-shot model on the base dataset and calculate the FP ratio of novel categories. The FP ratio of FSCE is up to 20$\%$, and the FP ratio of DeFRCN is up to 60$\%$, which are prominent and can not be overlooked. We think that current few-shot models have high FP ratio because of the small size of training dataset. A weak knowledge is formed in the learned model, which can only recognize the key features of an object. However, the key features of two categories may be the same, such as the sofa and chair both have the backrest. This will lead to high FP ratio. To reduce the FP, we wish to introduce the direct comparison between the TP and FP into the training phase. To this end, one can simply introduce the base images that contain the FP into the novel dataset. However, directly introducing many base images causes the badly data-imbalance issue, and makes the base categories dominate the training process.

In this work, we adopt cutmix~\cite{yun2019cutmix}, an image-mixing data augmentation approach. The naive use of cutmix ignores the FP of novel categories and causes the over-fitting problem in favor of the base categories, as
it generates more base samples than novel ones due to the fewer novel categories~\cite{yan2019meta}. Our key idea is to combine the novel images and the base images that contain the FP of novel categories. 
Specifically, we can Crop the FP region in the Base image and Paste it on a Novel image (CBPN), or we can Crop the Novel instance and Paste it on a Base image which contains FP (CNPB). CBPN uses a few novel images as the background which has the over-fitting problem, since the novel instances are repeated with the same image contexts. CNPB is superior because it can over-sample novel categories and leverage different FP images as the background. As illustrated in Figure~\ref{fig0}, CNPB can decrease the FP ratio of all methods up to 20$\%$. In CNPB, we investigate two key questions: (1) How to select useful base images?  and (2) How to combine novel and base data?

{\bf A multi-step selection strategy for base images} is proposed for our CNPB. First, we select the base images containing the FP of novel categories. Second, 
a small part of FP images are randomly selected to prevent the data-imbalance between the base and novel categories. Third, the base images containing the unlabeled ground truth or the FP that is easily confused with the novel instances (hard cases) are removed by CLIP~\cite{radford2021learning}. CLIP is a zero-shot recognition model which can use natural language to retrieve related images based on the CLIP score. 
We remove the hard cases of novel instances because we find using these cases reduces the accuracy of novel categories due to the weak knowledge of the few-shot model. For example, some special chairs are very similar to the sofa. If we use these chairs as negative samples, this will destroy the learned weak knowledge of the model, such as the backrest or the feet of sofa. 
At last, in order to balance TP and FP in each training iteration, the same category (SC) strategy is proposed. For example, the novel instance "cow" is pasted on the selected base image containing the FP of "cow". 
{\bf When combining the novel instances and the selected base images}, the novel instances are over-sampled by cropping and random down-sampling. We wish to avoid the base instances from being occluded. By randomly generating many candidate locations in a selected image, we can paste a processed novel instance at the assigned optimal location, which is determined by the minimum Intersection Over Union (IOU) between the candidates and the bounding boxes of original base instances and the FP regions in the selected base image. 

Our key contributions can be summarized as: (1) We propose CNPB, a novel data augmentation pipeline that Crops the Novel instances and Pastes them on the selected Base images, for FSOD. 
(2) Our method is simple yet effective, and can be easily integrated into existing FSOD methods without any architectural changes or complex algorithms. (3) Our CNPB significantly improves multiple baselines and achieves state-of-the-art performance on PASCAL VOC and MS COCO. 


\begin{figure*}[t]
	\begin{center}
		\centerline{\includegraphics[width=16cm]{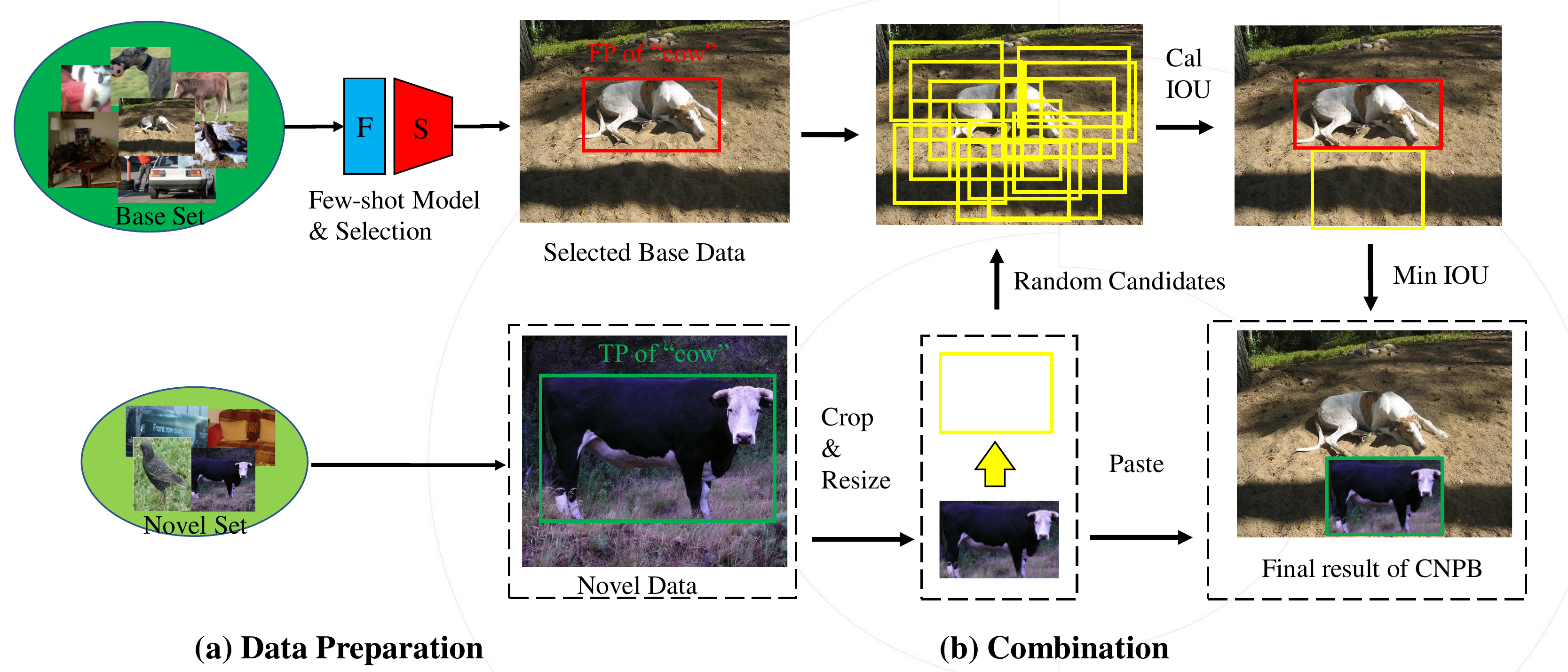}}
		\caption{We use a simple crop and paste method to build our CNPB pipeline. First, the base set are processed by the trained few-shot model. Then, the muti-step selection strategy is applied for selecting the useful base images which contain the FP of the novel category. After that, the novel instance from the novel set is combined with the selected base image. At last, the location of pasting the novel instance into the base image is decided by the minimum IOU between the random candidates and the FP /base instances in the selected base image.}
		\label{fig2}
	\end{center}
	\vskip -0.4in
\end{figure*}

\section{Related Works}
\subsection{Few-shot Object Detection}

FSOD is an important yet unsolved task in computer vision. Some works use meta-learning~\cite{fan2020few,kang2019few,yan2019meta,wang2019meta,li2021beyond}, where a meta-learner is introduced to acquire class-agnostic meta-knowledge which is transferred to novel classes. These methods extract meta-knowledge from a set of auxiliary tasks via the episode-based strategy~\cite{vinyals2016matching}, where each episode contains $C$ classes and $K$ samples of each class, i.e., $C$-way $K$-shot. 
With the help of a meta learner that takes the support images as well as the bounding box annotations as inputs, the feature re-weighting modules are applied to a single-stage object detector (YOLOv2)~\cite{kang2019few} and a two-stage object detector (Faster R-CNN)~\cite{yan2019meta}. 
CME~\cite{li2021beyond} uses a class margin equilibrium (CME) approach, with the aim to optimize both feature space partition and novel class reconstruction in a systematic way. Transformation Invariant Principle (TIP)~\cite{li2021transformation} is proposed for various meta-learning models by introducing consistency regularization on predictions from the transformed images.
TFA~\cite{wang2020frustratingly} is a simple two-stage fine-tuning approach, which significantly outperforms the earlier meta-learning methods. 
FSCE~\cite{sun2021fsce} proposes a simple yet effective approach to learning contrastive-aware object proposal encodings that facilitate the classification of detected objects. FADI~\cite{cao2021few} uses a two-step fine-tuning framework via association and discrimination, which builds up a discriminative feature space for each novel class with two integral steps. 
DeFRCN~\cite{qiao2021defrcn} extends Faster
R-CNN by using gradient decoupled layer for multi-stage decoupling and prototypical calibration block for multi-task decoupling. 

There are few FSOD methods are data-related.
MPSR~\cite{wu2020multi} adopts multi-scale positive sample refinement to handle scale variance problem. 
Pseudo-Labelling~\cite{kaul2022label} is proposed to 
obtain high-quality pseudo-annotations for novel categories from the training dataset. Their method can find previously unlabelled instances.
DetectorGAN~\cite{liu2019generative} uses a GAN~\cite{goodfellow2020generative} to generate new images, and jointly optimizes the GAN model and a detector. Different from them, we find the FP of the novel categories, and propose a new data-augmentation method for combining the novel instances and the base images which contain the FP of novel categories.

\subsection{Related Data Augmentation Techniques}

In computer vision, data augmentation strategies have been widely adopted, such as mixup~\cite{zhang2017mixup}, cutout~\cite{devries2017improved}, cutmix~\cite{yun2019cutmix} and mosaic~\cite{bochkovskiy2020yolov4}, etc. Among them, cutmix and mosaic can be used to do combination for novel and base categories in FSOD.
In cutmix~\cite{yun2019cutmix}, the patches are cut and pasted among training images where the ground truth labels are also mixed proportionally, which is used for image classification. Mosaic~\cite{bochkovskiy2020yolov4} mixes four training images into one image for small object detection. 
The most similar method to our CNPB pipeline is cutmix. However, there are several important differences. First,  we use the idea of cutmix to solve a new problem of FSOD. Second, the specific objects of cropping and pasting should be designed, in which we crop the novel instance and paste it on the selected base image. Furthermore, our selection strategies for base images are critical.
Without our designed strategies, existing copy-paste methods don't work in FSOD, such as mosaic in yolov4~\cite{bochkovskiy2020yolov4}.
There are also some recent efforts that employ cutmix to tackle the long-tail classification~\cite{park2022majority} and instance segmentation~\cite{ghiasi2021simple}. Unlike them, we consider the problem of FP by pasting a novel instance on the base image containing the FP and leverage a multi-step selection strategy for the base images in FSOD.


\section{Method}

\subsection{Preliminary}
In FSOD, given a labeled base dataset $D_B$, there are $C_B$ base classes with sufficient images in each class. Novel dataset $D_N$ with novel classes $C_N$ consists of a few samples in each class. $C_B$ and $C_N$ do not have overlapping categories. The number of objects for each class in $C_N$ is $K$ for K-shot detection. There are two stages in FSOD methods~\cite{wang2020frustratingly,sun2021fsce,li2021beyond}. In the pre-training stage, the model is trained on base classes to obtain a robust feature representation. In the fine-tuning stage, the pre-trained model is then fine-tuned on a balanced few-shot set which includes both base and novel classes ($C_B \cup C_N$).

\subsection{Our Pipeline}

We propose CNPB, a novel data augmentation pipeline that Crops the Novel instances and Pastes them on the selected Base images. 
There are two main steps in our CNPB.

{\bf Step1: Data Preparation.} For {\bf base data}, first, the fine-tuned few-shot model is used to test the base dataset, then the inference results are obtained. We select the base images as the background by using a multi-step selection strategy. In detail, the base images whose results have the FP of novel instances with higher confidence scores than the threshold are chosen. For each base category, we randomly select a certain amount (3 in our CNPB) of base samples from above FP images for all few-shot settings, indicating great transferable ability. After that, 
the base images that contain the unlabeled ground truth or the FP that is easily confused with the novel instances (hard cases) are removed by CLIP~\cite{radford2021learning}. 
CLIP is a zero-shot recognition model which can utilize natural language to retrieve related images based on the CLIP score. CLIP gets the input of a textual description of a novel category and an image, and then computes the distance between them. The template of the input text is "a" with the name of a novel category. The input image is the cropped instance which is predicted to have the novel category by the few-shot model in the base dataset. After inputting the text and current instance into the CLIP model, we can obtain the scores of all novel categories for the current image. If the maximum score is more than 0.5, this image is identified as a bad case which should be removed.  
Some samples removed by CLIP are shown in Figure~\ref{fig:vis} (a)/(b) in section~\ref{sec:visualization}.
Finally, to balance TP and FP in each training iteration, we select the base image that has the FP of the current pasted novel instance, which is a strategy of same category (SC). For example, the novel instance "cow" is pasted on the selected base image containing the FP of "cow".
For {\bf novel data}, the novel instances come from the few-shot novel set.

{\bf Step2: Combination.} We take "cow" as an example. 
Figure~\ref{fig2} shows how to combine the novel instance and the selected base image. Specifically, we crop and randomly downsize a novel instance $I_n$ of category $n$, and paste it on a selected base image. The optimal location for pasting is determined by the minimum IOU between the randomly generated candidates and the bounding boxes of original base instances $B_o=[b_{o1},b_{o2},..]$ and the FP regions $B_f=[b_{f1},b_{f2},..]$. The considered boxes $B=B_o \cup B_f$. If there are more than one considered boxes, the final IOU is calculated via summation. After getting the final location for pasting a novel instance, a new bounding box can be obtained. The formulation of this process is depicted below:

\begin{equation}
S = argmin(IOU(G(CD(I_n)),B))
\end{equation}
$CD$ denotes cropping and down-sampling for the novel instance $I_n$. $G$ is generating random candidates for pasting the novel instance. $IOU$ calculates the IOU between all candidates and the considered bounding boxes in a base image. $S$ is the final selected location.

{\bf The detail of combination.} In the step2 of CNPB, we design a majority-based and a minority-based combination methods, since the number of novel images and the number of the select base images may be different. Majority-based combination is duplicating the images, but minority-based combination means deleting the redundant images. For example, if the novel data is 10-shot and the number of the selected base images is 3, majority-based combination duplicates the base images to get 10 base images for pasting each novel instance on each base image. Minority-based combination deletes 7 novel images for pasting the remaining 3 novel instances on the 3 base images. Whether using majority-based combination or  minority-based combination, the diversity of novel categories does not change. Because the combined images will be added to the original few-shot set to build a new dataset for fine-tuning. The experimental results show that minority-based combination is better. The reason is that majority-based combination duplicates the base images which are the background of the pasted novel instances. The context with low diversity causes over-fitting problem.


\section{Experiments}

\subsection{Datasets and Evaluation Protocols}

We evaluate our methods on PASCAL VOC~\cite{everingham2010pascal,everingham2015pascal} and MS COCO~\cite{lin2014microsoft}. In PASCAL VOC, we adopt the common strategy~\cite{ren2016faster,redmon2017yolo9000}  that using VOC 2007 test set for evaluating while VOC 2007 and 2012 train/val sets are used for training. Following~\cite{yan2019meta}, 5 out of its 20 object categories are selected as the novel classes, while the remaining
as the base classes. We evaluate with three different novel/base splits from~\cite{yan2019meta}, named as split 1, split 2 and split 3.
Each split contains 15 base categories with abundant data and 5 novel categories with K annotated instances for K = 1, 3, 5, 10. Following~\cite{yan2019meta,wang2020frustratingly,sun2021fsce},  we use the mean average precision (mAP) of novel categories at 0.5 IoU threshold as the evaluation metric and report the results on the official test set of VOC 2007. When using MS COCO, 20 out of 80 categories are reserved as novel classes, the rest 60 categories are used as base classes. The detection performance with COCO-style AP, AP$_{50}$, and AP$_{75}$ for K = 10 and 30 shots of novel categories are reported.

\setlength{\tabcolsep}{8.5pt}
\begin{table*}[t]
\scriptsize

\begin{center}
	\caption{Comparison with state-of-the-art few-shot object detection methods on VOC2007 test set for novel classes of the three splits. {\bf black} indicates state-of-the-art. {\color{red}Red} is the improvement compared to the baseline. CNPB-TFA means applying our CNPB to TFA.}
	\label{table:1}
	\begin{tabular}{llcccccccccccc}
		\hline\noalign{\smallskip}
		&&\multicolumn{4}{c}{split 1} & 	\multicolumn{4}{c}{split 2}&\multicolumn{4}{c}{split 3}\\
		\hline\noalign{\smallskip}
		\multicolumn{2}{c}{Methods / Shots}& 1 & 3 &5& 10 & 1 &3&5&10&1&3&5&10\\
		\noalign{\smallskip}
		\hline
		\noalign{\smallskip}
		FRCN+ft~\cite{yan2019meta}&ICCV2019&11.9&29&36.9&36.9&5.9&23.4&29.1&28.8&5.0&18.1&30.8&43.4\\
		FRCN+ft-full~\cite{yan2019meta}&ICCV2019&13.8&32.8&41.5&45.6&7.9&26.2&31.6&39.1&9.8&19.1&35&45.1\\
		FR~\cite{kang2019few}&ICCV2019&14.8&26.7&33.9&47.2&15.7&22.7&30.1&40.5&21.3&28.4&42.8&45.9\\
		MetaDet~\cite{wang2019meta}&ICCV2019&18.9&30.2&36.8&49.6&21.8&27.8&31.7&43&20.6&29.4&43.9&44.1\\
		Meta R-CNN~\cite{yan2019meta}&ICCV2019&19.9&35&45.7&51.5&10.4&29.6&34.8&45.4&14.3&27.5&41.2&48.1\\
		TFA~\cite{wang2020frustratingly}&ICML2020&39.8&44.7&55.7&56.0&23.5 &34.1 &35.1 &39.1& 30.8 & 42.8& 49.5 &49.8\\
		MPSR~\cite{wu2020multi}&ECCV2020&41.7&51.4&55.2&61.8&24.4&39.2&39.9&47.8&35.6&42.3&48&49.7\\
		CME~\cite{li2021beyond}&CVPR2021&41.5&50.4&58.2&60.9&27.2&41.4&42.5&46.8&34.3&45.1&48.3&51.5\\
		FSCN~\cite{li2021few}&CVPR2021&40.7&46.5&57.4&62.4&27.3&40.8&42.7&46.3&31.2&43.7&50.1&55.6\\	
		HallucFsDet~\cite{zhang2021hallucination}&CVPR2021&47&46.5&54.7&54.7&26.3&37.4&37.4&41.2&40.4&43.3&51.4&49.6\\
		FSCE~\cite{sun2021fsce}&CVPR2021 &44.2&51.4&61.9&63.4&27.3&43.5&44.2&50.2&37.2&47.5&54.6&58.5\\
		UPE~\cite{wu2021universal}&ICCV2021&43.8&50.3 &55.4& 61.7& 31.2 & 41.2&42.2 & 48.3& 35.5 & 43.9 &50.6& 53.5 \\
		QA-FewDet~\cite{han2021query}&ICCV2021&42.4&55.7&62.6&63.4&25.9&46.6&48.9&51.1&35.2&47.8&54.8&53.5\\
		Meta faster-rcnn~\cite{han2021meta}&AAAI2021&43&60.6&66.1&65.4&27.7&46.1&47.8&51.4&40.6&53.4&{\bf59.9}&58.6\\
		FADI~\cite{cao2021few}&NIPS2021&50.3&54.2&59.3&63.2&30.6&40.3&42.8&48&45.7&49.1&55&59.6\\	
		DeFRCN~\cite{qiao2021defrcn}&ICCV2021&53.6&61.5&64.1&60.8&30.1&47.0&53.3&47.9&48.4&52.3&54.9&57.4\\
		FCT~\cite{han2022few}&CVPR2022&49.9&57.9&63.2&{\bf67.1}&27.6&43.7&49.2&51.2&39.5&52.3&57&58.7\\
		KFSOD~\cite{zhang2022kernelized}&CVPR2022&44.6&54.4&60.9&65.8&37.8&43.1&48.1&50.4&34.8&44.1&52.7&53.9\\	
		Pseudo-Labelling~\cite{kaul2022label}&CVPR2022&54.5& 58.8& 63.2& 65.7& 32.8& 50.7& 49.8& 50.6& 48.4 & 55.0& 59.6 &59.6\\
		\noalign{\smallskip}
		\hline
		\noalign{\smallskip}
		\multirow{2}*{CNPB-TFA}&Ours&48&52.8&58.9&59.1&25.9&42.1&38.7&43&35.4&44.1&50.5&52.3\\
		&Improve&{\color{red}+8.2}&{\color{red}+8.1}&{\color{red}+3.2}&{\color{red}+2.9}&{\color{red}+2.4}&{\color{red}+8}&{\color{red}+3.6}&{\color{red}+3.9}&{\color{red}+4.6}&{\color{red}+1.3}&{\color{red}+1}&{\color{red}+2.5}\\
		\multirow{2}*{CNPB-FSCE}&Ours&50.9&54.4&62.4&63&32.4&46.4&49.6&{\bf53.5}&40.1&48&53.4&57.8\\
		&Improve&{\color{red}+6.7}&{\color{red}+3}&{\color{red}+0.5}&-0.4&{\color{red}+5.1}&{\color{red}+2.9}&{\color{red}+5.4}&{\color{red}+3.3}&{\color{red}+2.9}&{\color{red}+0.5}&-1.2&-0.7\\			\multirow{2}*{CNPB-DeFRCN}&Ours&{\bf57.2}&{\bf63}&{\bf66.2}&66.6&{\bf39.7}&{\bf51.8}&{\bf54.7}&53.1&{\bf51}&{\bf56.9}&57&{\bf60.7}\\
		&Improve&{\color{red}+3.6}&{\color{red}+1.5}&{\color{red}+2.1}&{\color{red}+5.8}&{\color{red}+9.6}&{\color{red}+4.8}&{\color{red}+1.4}&{\color{red}+5.2}&{\color{red}+2.6}&{\color{red}+4.6}&{\color{red}+2.1}&{\color{red}+3.3}\\
		\hline	
	\end{tabular}
\end{center}
\vskip -0.2in
\end{table*}

\setlength{\tabcolsep}{10pt}	
\begin{table*}[t]
\scriptsize
\begin{center}
	\caption{Few-shot object detection performance on MS COCO. {\bf Black} indicates the state-of-the-art. {\color{red}Red} is the improvement compared to the baseline. CNPB-TFA means applying our CNPB to TFA.}
	\label{table:2}
	\begin{tabular}{lcccccc}
		\hline\noalign{\smallskip}
		&\multicolumn{2}{c}{novel AP}&\multicolumn{2}{c}{novel AP$_{50}$}&	\multicolumn{2}{c}{novel AP$_{75}$}\\
		\hline\noalign{\smallskip}
		Methods / Shots& 10 & 30 & 10 & 30 & 10 & 30 \\
		\noalign{\smallskip}
		\hline
		\noalign{\smallskip}
		FR~\cite{kang2019few}&5.6&9.1&12.3&19&4.6&7.6\\
		Meta R-CNN~\cite{yan2019meta} &8.7&12.4&19.1&25.3&6.6&10.8\\	
		TFA~\cite{wang2020frustratingly}&10&13.7&-&-&9.3&13.4\\
		MSPR~\cite{wu2020multi}&9.8&14.1&17.9&25.4&9.7&14.2\\
		CME~\cite{li2021beyond}&15.1&16.9&24.6&28&16.4&17.8\\
		FSCN~\cite{li2021few}&11.3&15.1&20.3&29.4&-&-\\
		FSCE~\cite{sun2021fsce}&11.1&15.3&-&-&9.8&14.2\\
		UPE~\cite{wu2021universal}&11&15.6&-&-&10.7&15.7\\
		QA-FewDet~\cite{han2021query}&10.2&11.5&20.4&23.4&9.0&10.3\\
		Meta faster-rcnn~\cite{han2021meta}&12.7&16.6&25.7&31.8&10.8&15.8\\
		FADI~\cite{cao2021few}&12.2&16.1&22.7&29.1&11.9&15.8\\
		DeFRCN~\cite{qiao2021defrcn}&18.5&22.6&-&-&-&-\\
		FCT~\cite{han2022few}&17.1&21.4&-&-&-&-\\
		KFSOD~\cite{zhang2022kernelized}&18.5&-&26.3&-&18.7&-\\
		Pseudo-Labelling~\cite{kaul2022label}&17.8 &{\bf24.5} &30.9 &{\bf41.1}& 17.8& {\bf25.0}\\
		\noalign{\smallskip}
		\hline
		\noalign{\smallskip}			
		CNPB-TFA&10.6 ({\color{red}+0.6})&15 ({\color{red}+1.3})&20.7 (-)&27.5 (-)&9.8 ({\color{red}+0.5})&14.4 ({\color{red}+1})\\
		CNPB-FSCE&11.6 ({\color{red}+0.5})&16 ({\color{red}+0.7})&23.7 (-)&30.9 (-)&10.2 ({\color{red}+0.4})&14.9 ({\color{red}+0.7})\\
		CNPB-DeFRCN&{\bf20.3} ({\color{red}+1.8})&23.1 ({\color{red}+0.5})&{\bf35.5} (-)&40 (-)&{\bf20.3} (-)&23.8 (-)\\
		\hline				
	\end{tabular}
\end{center}
\vskip -0.2in
\end{table*}

\subsection{Implementation Details}
Our baselines are TFA~\cite{wang2020frustratingly}, FSCE~\cite{sun2021fsce} and DeFRCN~\cite{qiao2021defrcn}, which are representative methods in FSOD. These methods all use Faster R-CNN~\cite{ren2016faster} with  ResNet-101~\cite{he2016deep}, which are the same as almost all FSOD methods. The training strategies of our methods follows the selected baselines. The difference is that we pre-process the few-shot set using our CNPB pipeline. 
There are some hyper-parameters in our pipeline. 
The threshold for selected base images is 0.6 for TFA and DeFRCN, and 0.8 for FSCE. Resize ratio of the novel instances is randomly sampled from 1/5 to 1/2. The number of candidates for pasting the novel instances on a base image is 1000.

\subsection{Comparison with State-of-the-art Methods}

We compare our approach to several competitive FSOD methods. The results are shown in Table~\ref{table:1} and Table~\ref{table:2}. We adopt our CNPB for all baselines to prove the effectiveness of our pipeline. 
Following~\cite{han2021meta,li2021beyond,wu2020multi,cao2021few,kaul2022label}, we use a single run with the same training images to get the results of different shots.

\subsubsection{Results on PASCAL VOC and MS COCO.} 
Following~\cite{yan2019meta,wang2020frustratingly,sun2021fsce}, we provide the AP$_{50}$ of the novel classes on PASCAL VOC with three splits in Table~\ref{table:1}. By using CNPB, our methods all can outperform the baselines in almost all few-shot settings and achieve state-of-the-art performance. The most obvious improvement is up to 9.6$\%$. 

We report the COCO-style AP, AP$_{50}$, and AP$_{75}$ of the 20 novel classes on MS COCO in Table~\ref{table:2}. By using CNPB, our method sets a new state-of-the-art record for 10-shot setting, exceeding the second method (Pseudo-Labelling) by 3$\%$ on average. The performance of our method on 30 shot setting is only 1$\%$ lower than the best performance. Therefore, the overall performance of our method is the best. However, the improvement is not as significant as that of PASCAL VOC, since there are higher shots in the setting of MS COCO, which declines the influence of the combined images.

\subsection{Ablation Study}

Following~\cite{yan2019meta,sun2021fsce}, we do ablation studies on PASCAL VOC split 1. In most cases, we use FSCE as the baseline.

\subsubsection{Trained with CNPB}

There are pre-training and fine-tuning steps in FSOD. CNPB is used to train a new model based on the fine-tuned model from the fine-tuning stage. 
The training strategies when using CNPB are the same as that of the fine-tuning step. 
In Table~\ref{table:3}, using CNPB can achieve the best performance. The results of training for three steps without CNPB (+ft) show that fine-tuning more iterations can not bring obvious improvement. 
In Table~\ref{table:3}, we also see that using CNPB can not hurt the performance of base classes, and even achieves some improvement in most of the settings.


\begin{table}[t]
\scriptsize
\setlength{\tabcolsep}{6pt}		
\begin{center}
	\caption{The results of using CNPB or not. The AP$_{50}$ of novel and base categories (separated by /) of TFA and FSCE are reported. ft is fine-tuning the model without CNPB.}
	\label{table:3}
	\begin{tabular}{lcccc}
		\hline\noalign{\smallskip}
		Methods / Shots & 1 & 3 &5& 10 \\
		\noalign{\smallskip}
		\hline
		\noalign{\smallskip}
		TFA (Our impl.)~\cite{wang2020frustratingly}&40/79.4&47.7/79.4&56/79.3&58.1/79.5\\
		TFA+ft&41.9/79.4&48/79.2&55.7/79.1&58.2/79.2\\
		TFA+CNPB&{\bf48}/79&{\bf52.8}/79.4&{\bf58.9}/79.4&{\bf59.1}/79.4\\
		\hline
		\noalign{\smallskip}
		FSCE (Our impl.)~\cite{sun2021fsce}&44.7/78.3&50.2/75.4&59/75.8&61.7/75.7\\
		FSCE+ft&43.5/77.5&49.9/74.9&59.4/74.3&61.7/75.5\\
		FSCE+CNPB&{\bf50.9}/77.9&{\bf54.4}/76.7&{\bf62.4}/76&{\bf63}/76.8\\
		\hline
	\end{tabular}
\end{center}
\vskip -0.2in
\end{table}

\begin{table}[t]
\scriptsize
\setlength{\tabcolsep}{11pt}	
\begin{center}
	\caption{The necessary of our selection strategies. {\em w/o} FP or {\em w/o} SC means the novel instance can be pasted to the random base image or the random FP image of the selected base image, respectively. CNPB-n means CNPB with n selected base image(s). - means the results are equal to that of the upper row. {\em w/o} remove does not remove the bad cases.}
	\label{table:6}
	\begin{tabular}{lcccc}
		\hline\noalign{\smallskip}
		Setting / Shot& 1 & 3 &5& 10 \\
		\noalign{\smallskip}
		\hline
		\noalign{\smallskip}
		FSCE(Our impl.)~\cite{sun2021fsce}&44.7&	50.2&	59&	61.7\\
		FSCE+ft&43.5&49.9&59.4&61.7\\
		\hline
		\noalign{\smallskip}
		CNPB {\em w/o} FP &45.4&51.4&58.2&62.2\\
		CNPB {\em w/o} SC &47.5&53.2&61.3&61.9\\		CNPB&{\bf50.9}&{\bf54.4}&{\bf62.4}&{\bf63}\\
		\hline
		\noalign{\smallskip}
		CNPB-1&{\bf46.8}&{\bf54.3}&59.3&62.1\\
		CNPB-3&-&49.6&{\bf59.7}&{\bf62.5}\\
		CNPB-5&-&-&58.4&62\\
		CNPB-10&-&-&-&60.6\\
		\hline
		\noalign{\smallskip}
		Majority-based&47.5&	53.4	&61	&61.1\\
		Minority-based&{\bf50.9}&{\bf54.4}&{\bf62.4}&{\bf63}\\
		
		\hline
		\noalign{\smallskip}			
		CNPB {\em w/o} remove &43.6&	50.7&61.1&61.5\\
		CNPB&{\bf50.9}&{\bf54.4}&{\bf62.4}&
		{\bf63}\\
		\hline
		\noalign{\smallskip}
	\end{tabular}
\end{center}
\vskip -0.2in
\end{table}

\subsubsection{The Multi-step Selection Strategy}

In Table~\ref{table:6}, we show the results without our selection strategies, including the novel instance is pasted on the base image without FP or the selected base image without the same category (SC), the different number of the select base images, and using the selected base images without removing the bad cases. 

First, compared to CNPB, CNPB without FP causes severe accuracy degradation, however it is still superior to the baseline. CNPB without SC is better than CNPB without FP, and CNPB achieves the best performance. The results demonstrate that the selected FP images is necessary, and applying SC can bring further improvement. Second, using a small number of selected base images, such as 3 or 1, can achieve better performance, because more base images with repeatedly novel samples may cause the over-fitting on the novel categories. Since we use the minority-based method, the results of one base with one novel (1-shot) are equal to more base with one novel. Given 3 selected base images, minority-based is better than majority-based in all settings. However, majority-based is still better than the baseline, which means our CNPB is robust to the combination methods. Third, CNPB without removing can bring slight improvement than the baseline, but using remove operation can obtain much higher performance. CNPB-n uses the minority-based combination with threshold 0.7 to 0.8 for higher performance without removing operation. The threshold of other experiments is 0.8. 

\subsubsection{The Methods of Combination}
\label{sec:method_combination}

We study different types of combination methods for our few-shot setting in Table~\ref{table:5}. Addition is no combination at the image level, just puts all selected base images into the novel dataset for fine-tuning. Addition {\em w/o sel} means without using our selection strategy and directly using all FP images whose confidence scores on a novel category are higher than 0.8. Mosaic~\cite{bochkovskiy2020yolov4} uses one novel image and three selected base images to obtain a final image in our case. CP is using the original setting of cutmix for our tasks, in which the novel and base images are randomly selected. CBPN crops the FP regions in the selected base images and pastes them on the novel images. CNPB crops the novel instances and pastes them on the selected base images.
In Table~\ref{table:5}, our CNPB can obtain the best performance in almost all settings. Addition without selection shows a poor performance, and Addition with selection can not improve the accuracy. Using mosaic can bring modest improvement. CP is better than CBPN, which uses few novel images as the context, causing the over-fitting on these novel images. 

\begin{table}[t]
\scriptsize
\setlength{\tabcolsep}{12pt}		
\begin{center}
\caption{The methods of combination based on FSCE. All methods are based on FSCE+ft.}
\label{table:5}
\begin{tabular}{lcccc}
	\hline\noalign{\smallskip}
	Setting / Shot& 1 & 3 &5& 10 \\
	\noalign{\smallskip}
	\hline
	\noalign{\smallskip}
	FSCE(Our impl.)~\cite{sun2021fsce}&44.7&	50.2&	59&	61.7\\
	FSCE+ft&43.5&49.9&59.4&61.7\\
	+ Addition {\em w/o sel}&37.7&44.5&55.2&58.2\\
	+ Addition&43.5&	48.7&60.2&	61.3\\
	+ Mosaic~\cite{bochkovskiy2020yolov4}&44.4&50.9&59.1&{\bf63}\\
	+ CP&44.2&48&59&61.5\\
	+ CBPN&43.9&	47.1&	58.6&	61.2\\
	+ CNPB&{\bf50.9}&{\bf54.4}&{\bf62.4}&{\bf63}\\
	\hline
\end{tabular}
\end{center}
\vskip -0.2in
\end{table}

\begin{table}[t]
\scriptsize
\setlength{\tabcolsep}{7pt}		
\begin{center}
\caption{Factors in our pipeline. We select different intervals for threshold.  Data augmentation uses several common methods. The number of novel candidates is from 10 to 10000. The resize ratios of down-sampling include random times (generated from 2 to 5) or 2 times or 5 times. All experiments remove the bad cases. }
\label{table:8}
\begin{tabular}{llcccc}
	\hline\noalign{\smallskip}
	Factors &Setting / Shot& 1 & 3 &5& 10 \\
	\noalign{\smallskip}
	\hline
	\noalign{\smallskip}
	\multirow{4}*{Threshold}&0.5-0.6&46.3&52.4&60.3&{\bf63.8}\\
	&0.6-0.7&45.2&{\bf55}&60.9&62.8\\
	&0.7-0.8&45.8&52.3&60.6 &62.5\\
	&0.8-1&{\bf50.9}&54.4&{\bf62.4}&63\\
	\hline
	\noalign{\smallskip}
	\multirow{6}*{Data Aug}		
	&No&{\bf50.9}&{\bf54.4}&{\bf62.4}&{\bf63}\\
	&+ Random Crop&51	&53.1	&61.8	&62.9\\
	&+ Colorjitter&{\bf50.9}&	53.5&	61.3&62.7\\
	&+ Hflip&49.5	&52.3&	61.8&62.2\\
	&+ Random Rot&50.7&52&61.3&62.6\\
	&+ TrivialAug~\cite{muller2021trivialaugment}& 49.4&52.8&61.7	&62.2\\
	\hline
	\noalign{\smallskip}
	\multirow{4}*{Novel Candidates}&10&47.7&52.9&	62.3&62.9\\
	&100&50.6&53.5&	61.2&62.6\\
	&1000&{\bf50.9}&{\bf54.4}&{\bf62.4}&63\\
	&10000&50.4&	53.6&61.2&{\bf63.7}\\
	\hline
	\noalign{\smallskip}
	\multirow{4}*{Resize Ratio}&2 to 5&{\bf50.9}&{\bf54.4}&{\bf62.4}&63\\
	&2&49.5&{\bf54.4}&61.5	&61.8\\
	&5&49.8&53.3&61.9&{\bf63.4}\\			
	\hline
\end{tabular}
\end{center}
\vskip -0.2in
\end{table}

\begin{figure*}[t]
	\centering
	\includegraphics[width=0.8\linewidth]{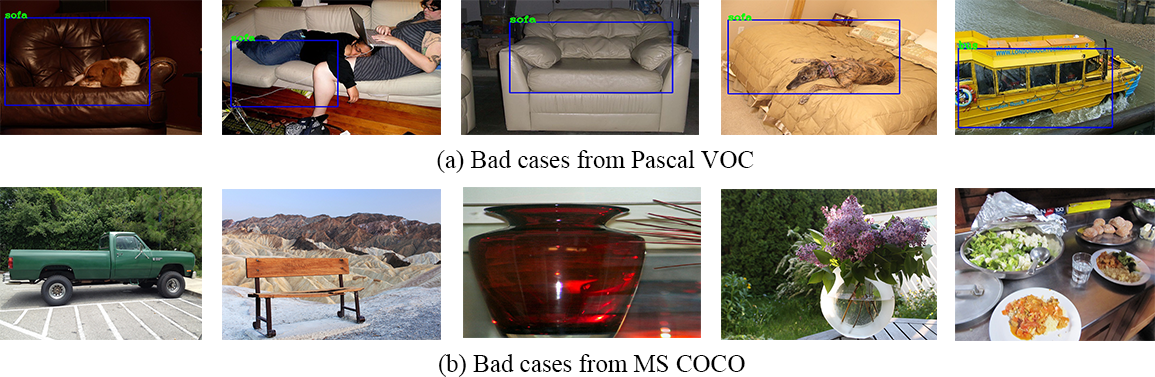}
	
	\caption{Bad cases removed by CLIP in our pipeline. (a) Bad cases from PASCAL VOC include some unlabeled ground truths and confusing hard cases for the few-shot model. (b) is the bad cases removed by CLIP. These are all novel categories of MS COCO.}
	\label{fig:vis}
	\vskip -0.2in
\end{figure*}

\subsubsection{Other Factors}
In Table~\ref{table:8}, we explore the influence of some important factors in our CNPB pipeline, such as threshold, data augmentation, the number of novel candidates, and resize ratio. Threshold is used to select the base image whose predicted score of a novel category is higher than the threshold. Data augmentation is applying common data augmentation methods on pasted novel instances. The number of novel candidates means the number of randomly generated locations for novel instances. Resize ratio is the downsize ratio of the novel instance. To sum up, setting the threshold as 0.8 to 1 achieves relatively better results.  Using data augmentation on novel instances is not useful. The best number of novel candidates is 1000. The random resize ratio is better than a fixed ratio.

\subsubsection{Visualization}
\label{sec:visualization}

The visualization of some bad cases removed by CLIP is shown in Figure~\ref{fig:vis}.
(a) Bad cases from PASCAL VOC include some unlabeled ground truth, such as the sofa in the first two images. The rest images of (a) are some hard cases for the few-shot model. For example, the chair/bed is like a small/large version of sofa, and the last boat has the appearance of bus. These targets can hurt the performance of few-shot model, since it just starts to have the ability to recognize these obvious features.
(b) is some bad cases in MS COCO removed by CLIP, such as car, chair, bottle, plotted plant, and dining table. These are all unlabeled novel categories of MS COCO.

\section{Conclusion}

In this paper, we propose a novel data augmentation pipeline, called CNPB, that crops the novel instances and pastes them on the selected base images. In detail, we design a multi-step strategy to select the base images. Our method can be easily integrated into existing FSOD methods. Extensive experiments on the few-shot object detection datasets, \ie, PASCAL VOC and MS COCO, validate the effectiveness of our method.  
\section*{Acknowledgement} This work is sponsored Hetao Shenzhen-Hong Kong Science and Technology Innovation Cooperation Zone (HZQB-KCZYZ-2021045).

{\small
\bibliographystyle{ieee_fullname}
\bibliography{egbib}
}

\end{document}